\documentclass[lettersize,journal]{IEEEtran}
\usepackage{amsmath,amsfonts,amssymb}
\usepackage{algorithmic}
\usepackage{algorithm}
\usepackage{array}
\usepackage[caption=false,font=normalsize,labelfont=sf,textfont=sf]{subfig}
\usepackage{textcomp}
\usepackage{stfloats}
\usepackage{url}
\usepackage{verbatim}
\usepackage{graphicx}
\usepackage{multirow}
\usepackage{cite}
\usepackage{multicol}
\begin{document}

\title{VeloxNet: Efficient Spatial Gating for Lightweight Embedded Image Classification}

\author{Md Meftahul Ferdaus,
        Elias Ioup,
        Mahdi Abdelguerfi,
        Anton Netchaev,
        Steven Sloan,
        Ken Pathak,
        and~Kendall N. Niles
        \thanks{M. Ferdaus and M. Abdelguerfi are with the Canizaro Livingston Gulf States Center for Environmental Informatics, the University of New Orleans, New Orleans, USA (e-mail: mferdaus@uno.edu; gulfsceidirector@uno.edu).}
        \thanks{E. Ioup is with the Center for Geospatial Sciences, Naval Research Laboratory, Stennis Space Center, Hancock County, Mississippi, USA.}
        \thanks{A. Netchaev, S. Sloan, K. Pathak, and K. N. Niles are with the US Army Corps of Engineers, Engineer Research and Development Center, Vicksburg, Mississippi, USA.}
        \thanks{This work has been submitted to the IEEE for possible publication. Copyright may be transferred without notice, after which this version may no longer be accessible.}
        }



\maketitle

\begin{abstract}
Deploying deep learning models on embedded devices for tasks such as aerial disaster monitoring and infrastructure inspection requires architectures that balance accuracy with strict constraints on model size, memory, and latency. This paper introduces VeloxNet, a lightweight CNN architecture that replaces SqueezeNet's fire modules with gated multi-layer perceptron (gMLP) blocks for embedded image classification. Each gMLP block uses a spatial gating unit (SGU) that applies learned spatial projections and multiplicative gating, enabling the network to capture spatial dependencies across the full feature map in a single layer. Unlike fire modules, which are limited to local receptive fields defined by small convolutional kernels, the SGU provides global spatial modeling at each layer with fewer parameters. We evaluate VeloxNet on three aerial image datasets: the Aerial Image Database for Emergency Response (AIDER), the Comprehensive Disaster Dataset (CDD), and the Levee Defect Dataset (LDD), comparing against eleven baselines including MobileNet variants, ShuffleNet, EfficientNet, and recent vision transformers. VeloxNet reduces the parameter count by 46.1\% relative to SqueezeNet (from 740,970 to 399,366) while improving weighted F1 scores by 6.32\% on AIDER, 30.83\% on CDD, and 2.51\% on LDD. These results demonstrate that substituting local convolutional modules with spatial gating blocks can improve both classification accuracy and parameter efficiency for resource-constrained deployment. The source code will be made publicly available upon acceptance of the paper.
\end{abstract}


\begin{table*}[t]
\centering
{\normalsize\textbf{Nomenclature}}\\[4pt]
{\footnotesize
\begin{tabular}{@{}p{0.08\textwidth} p{0.35\textwidth} | p{0.08\textwidth} p{0.35\textwidth}@{}}
\hline
\multicolumn{4}{@{}l}{\textbf{Acronyms}} \\[1pt]
AIDER & Aerial Image Database for Emergency Response & LDD & Levee Defect Dataset \\
CDD & Comprehensive Disaster Dataset & PTQ & Post-Training Quantization \\
CNN & Convolutional Neural Network & QAT & Quantization-Aware Training \\
ERM & Empirical Risk Minimization & SGU & Spatial Gating Unit \\
FLOPs & Floating-Point Operations & SVD & Singular Value Decomposition \\
FPS & Frames Per Second & NLP & Natural Language Processing \\
GeLU & Gaussian Error Linear Unit & & \\
gMLP & Gated Multi-Layer Perceptron & & \\[3pt]
\hline
\multicolumn{4}{@{}l}{\textbf{Symbols}} \\[1pt]
$\mathcal{D}$ & Labeled training set $\{(\mathbf{x}_i, y_i)\}_{i=1}^{N}$ & $\mathcal{L}$ & Cross-entropy loss function \\
$\mathbf{x}_i$ & Input image $\in \mathbb{R}^{H \times W \times C}$ & $\mathcal{F}$ & Fire module operation \\
$y_i$ & Class label $\in \{1, \dots, K\}$ & $L$ & Number of gMLP blocks \\
$H, W, C$ & Image height, width, and channels & $n$ & Number of spatial tokens \\
$K$ & Number of target classes & $d$ & Feature dimension (channel size) \\
$N$ & Number of training samples & $\mathbf{Z}$ & Intermediate representation $\in \mathbb{R}^{n \times d}$ \\
$f_\theta$ & Classifier parameterized by $\theta$ & $\mathbf{Z}_1, \mathbf{Z}_2$ & Channel-wise splits of $\mathbf{Z}$ \\
$\theta$ & Learnable model parameters & $\mathbf{W}_g$ & Spatial projection weights $\in \mathbb{R}^{n \times n}$ \\
$P(\theta)$ & Total trainable parameters & $\mathbf{b}_g$ & Spatial projection bias \\
$\mathcal{C}(f_\theta)$ & Computational cost (FLOPs) & $U, V$ & Channel projection matrices \\
$\mathcal{M}(\theta)$ & Model storage size (bytes) & $\sigma$ & GeLU activation function \\
$\mathcal{T}(f_\theta)$ & Inference latency & $\odot$ & Hadamard (element-wise) product \\
$\mathcal{A}(f_\theta)$ & Accuracy metric (e.g., F1 score) & $s$ & Squeeze ratio in fire modules \\
 & & $e_1, e_3$ & Expand dimensions in fire modules \\[1pt]
\hline
\end{tabular}}
\end{table*}

\section{Introduction}
\IEEEPARstart{E}{mbedded} devices used in applications such as drone-based disaster monitoring, mobile inspection systems, and infrastructure defect assessment \cite{alshawi2025imbalance, kuchi2019machine} impose tight constraints on model size, memory, and inference latency. Convolutional neural networks (CNNs) are effective for image classification, but deploying them on such hardware requires careful trade-offs between model capacity and computational cost. In practice, reducing model size usually comes at the expense of accuracy, and this trade-off has motivated a large body of work on efficient architectures.

Prior work on efficient deep learning falls broadly into two categories: post-training compression and efficient architecture design. Post-training compression techniques such as pruning, quantization, and Huffman encoding reduce the size and computational cost of pretrained networks. These methods induce sparsity or reduce numerical precision, making models smaller at the cost of some accuracy loss.

Pruning involves removing less important weights or neurons from the network, which can be executed either in a structured manner (removing entire filters or channels) or in an unstructured manner (removing individual weights). This approach helps reduce model size and computational load but traditionally leads to accuracy degradation if not carefully managed. For instance, \cite{han2015deep} introduced a method to prune networks by learning only the important connections, followed by retraining to fine-tune the remaining weights. Recent advancements demonstrate that pruning combined with quantization can achieve significant compression while minimizing accuracy loss \cite{tang2020automated, bai2023unified}.

Quantization reduces the number of bits required to represent each weight, typically transitioning from 32-bit floating-point to lower bit-width representations like 8-bit integers, thereby reducing model size and accelerating inference. Techniques such as quantization-aware training (QAT) and post-training quantization (PTQ) help maintain accuracy while reducing precision. \cite{wen2020exploiting} highlight the redundancy in CNN weight tensors that can be exploited to further improve compression ratios. \cite{kwon2019structured} propose a structured representation for sparse quantized weights, achieving high compression ratios with efficient decoding.

Huffman encoding, a lossless compression technique, further minimizes storage requirements by encoding frequently occurring weights with fewer bits, often used alongside pruning and quantization to maximize compression benefits. \cite{gajjala2020huffman} propose several Huffman-based encoding techniques that outperform traditional methods in reducing encoded data volume during distributed deep learning. \cite{oswald2018optimal} enhance Huffman encoding through frequent pattern mining, achieving superior compression ratios compared to conventional approaches.

Another line of work focuses on architectural techniques such as low-rank decomposition and lightweight model design. Low-rank decomposition factorizes weight matrices into lower-dimensional forms, reducing both parameter count and computation. \cite{swaminathan2020sparse} proposed a sparse low-rank method that sparsifies SVD matrices to achieve high compression with minimal accuracy loss. \cite{sobolev2022pars} introduced PARS, which uses Bayesian optimization to select optimal rank combinations for compression. \cite{phan2020stable} developed a stable low-rank tensor decomposition method for CNNs.

An alternative approach is to design compact architectures from scratch. SqueezeNet achieves AlexNet-level accuracy on ImageNet with 50$\times$ fewer parameters by using fire modules that compress and expand channels \cite{iandola2016squeezenet}. MobileNet uses depthwise separable convolutions to reduce parameters and FLOPs while maintaining competitive accuracy \cite{howard2017mobilenets}.

Further architectural innovations include techniques such as the EfficientNet model, which scales network depth, width, and resolution systematically using a compound scaling method, achieving state-of-the-art performance with fewer parameters and computations \cite{tan2019efficientnet}. Inception modules and residual connections have been explored to improve deep network optimization. Inception modules, introduced by Szegedy et al., enhance performance by increasing network width and capturing multi-scale features within the same layer \cite{szegedy2015going}. Residual connections, introduced by He et al., enable training of much deeper networks by allowing gradients to flow without vanishing, significantly improving training efficiency and accuracy \cite{he2016deep}. Combining these techniques, such as Inception-ResNet, has been shown to accelerate training and improve performance by leveraging the strengths of both architectures \cite{szegedy2016inception}.

Recent work on efficient embedded vision has also explored widened atrous convolutions with attention mechanisms to improve accuracy under resource constraints \cite{ferdaus2025widened}. More broadly, attention mechanisms have shown strong results for modeling long-range dependencies. The Transformer architecture \cite{vaswani2017attention} achieves strong performance across many tasks, but the quadratic cost of self-attention limits its use on embedded hardware. Liu et al. \cite{liu2021pay} introduced Gated MLPs (gMLPs), showing that spatial gating can replace attention in certain settings while reducing computational cost. Their work focused on large-scale vision and NLP tasks using models with millions of parameters.

However, several gaps remain. The original gMLP work targeted large-scale settings (ImageNet, CIFAR) with substantial compute budgets and did not explore integration with lightweight CNNs such as SqueezeNet or MobileNet. There is no prior work on adapting gMLP blocks for parameter-constrained embedded scenarios, and domain-specific applications like aerial imagery have not been evaluated.

We propose VeloxNet (from Latin \textit{velox}, meaning swift) to address these gaps by replacing SqueezeNet's fire modules with gMLP blocks designed for low-parameter settings. Our contributions are: (1) an architectural integration strategy that substitutes fire modules with gMLP blocks while preserving the overall SqueezeNet topology; (2) efficiency-oriented design choices including fixed 156-channel dimensions, simplified normalization, and compact spatial projections suited to embedded hardware; (3) evaluation on three aerial image datasets (AIDER, CDD, Levee Defect) not covered in prior gMLP work; and (4) a detailed comparison between fire modules and gMLP blocks in terms of parameter efficiency and spatial modeling.

The key idea is that gMLP blocks provide global spatial modeling through learned spatial projections, whereas fire modules are limited to local receptive fields. By making this substitution, VeloxNet achieves lower parameter counts than SqueezeNet while improving classification accuracy on all three datasets.

\section{Problem Formulation}

In this section, we formally define the problem of efficient image classification under resource constraints and establish the mathematical framework that motivates VeloxNet's architectural design. A summary of all acronyms and mathematical symbols used throughout this paper is provided in the Nomenclature. We adopt notation consistent with prior work on lightweight architectures \cite{iandola2016squeezenet, howard2017mobilenets} and gated MLPs \cite{liu2021pay}.

\subsection{Notation and Preliminaries}

Let $\mathcal{D} = \{(\mathbf{x}_i, y_i)\}_{i=1}^{N}$ denote a labeled training set, where $\mathbf{x}_i \in \mathbb{R}^{H \times W \times C}$ represents an input image of height $H$, width $W$, and $C$ channels, and $y_i \in \{1, \dots, K\}$ is the corresponding class label over $K$ categories. A neural network classifier $f_\theta: \mathbb{R}^{H \times W \times C} \rightarrow \Delta^{K-1}$ parameterized by $\theta \in \mathbb{R}^{P}$ maps each input to a probability simplex $\Delta^{K-1}$, where $P = |\theta|$ denotes the total number of trainable parameters.

We define the following quantities central to our formulation:

\begin{itemize}
    \item \textbf{Parameter count}: $P(\theta) = |\theta|$, the total number of scalar learnable parameters in the model.
    \item \textbf{Computational cost}: $\mathcal{C}(f_\theta)$, measured in floating-point operations (FLOPs) for a single forward pass on input $\mathbf{x} \in \mathbb{R}^{H \times W \times C}$.
    \item \textbf{Model size}: $\mathcal{M}(\theta)$, the storage footprint of the model in bytes under a given numerical precision.
    \item \textbf{Inference latency}: $\mathcal{T}(f_\theta)$, the wall-clock time for a single forward pass on a target hardware platform.
\end{itemize}

\subsection{Resource-Constrained Image Classification}

The standard empirical risk minimization (ERM) objective for image classification seeks parameters $\theta^*$ that minimize the expected loss:

\begin{equation}
\theta^* = \arg\min_{\theta} \; \mathbb{E}_{(\mathbf{x}, y) \sim \mathcal{D}} \left[ \mathcal{L}\big(f_\theta(\mathbf{x}), y\big) \right],
\label{eq:erm}
\end{equation}

where $\mathcal{L}$ denotes the cross-entropy loss. In the embedded vision setting, this optimization must additionally satisfy hard resource constraints imposed by the target deployment platform:

\begin{equation}
\begin{aligned}
\theta^* = \arg\min_{\theta} \quad & \mathbb{E}_{(\mathbf{x}, y) \sim \mathcal{D}} \left[ \mathcal{L}\big(f_\theta(\mathbf{x}), y\big) \right] \\
\text{subject to} \quad & P(\theta) \leq P_{\max}, \\
& \mathcal{C}(f_\theta) \leq C_{\max}, \\
& \mathcal{M}(\theta) \leq M_{\max}, \\
& \mathcal{T}(f_\theta) \leq T_{\max},
\end{aligned}
\label{eq:constrained_opt}
\end{equation}

where $P_{\max}$, $C_{\max}$, $M_{\max}$, and $T_{\max}$ are platform-specific budgets for parameters, computation, memory, and latency, respectively. In practice, most lightweight architectures address a subset of these constraints, often treating parameter count and FLOPs as primary optimization targets.

\subsection{The Efficiency-Accuracy Trade-off}

Existing lightweight architectures implicitly assume a monotonic trade-off between model capacity and resource consumption. Formally, let $\mathcal{A}(f_\theta)$ denote a task-specific accuracy metric (e.g., F1 score). The conventional Pareto frontier of lightweight CNN design is characterized by:

\begin{equation}
\frac{\partial \mathcal{A}}{\partial P} > 0 \quad \text{and} \quad \frac{\partial \mathcal{A}}{\partial \mathcal{C}} > 0,
\label{eq:tradeoff}
\end{equation}

implying that improving accuracy requires increasing either parameter count or computational cost. Architectures such as SqueezeNet \cite{iandola2016squeezenet} address this by designing parameter-efficient building blocks (fire modules), but remain constrained by the local receptive fields of convolutional operations. Specifically, a fire module $\mathcal{F}$ with squeeze ratio $s$ and expand dimensions $e_1, e_3$ operates as:

\begin{equation}
\mathcal{F}(\mathbf{h}) = \text{Concat}\big[\text{Conv}_{1\times1}^{e_1}(\mathbf{z}), \; \text{Conv}_{3\times3}^{e_3}(\mathbf{z})\big], \quad \mathbf{z} = \text{Conv}_{1\times1}^{s}(\mathbf{h}),
\label{eq:fire_module}
\end{equation}

where $\mathbf{h}$ is the input feature map. The aggressive channel compression through $s \ll e_1 + e_3$ introduces information bottlenecks, and the reliance on local convolutions limits the spatial modeling capacity to the kernel receptive field.

\subsection{Spatial Gating as an Alternative Inductive Bias}

Gated MLPs \cite{liu2021pay} introduce an alternative mechanism for spatial modeling through multiplicative gating. Given an intermediate representation $\mathbf{Z} \in \mathbb{R}^{n \times d}$ (where $n$ is the number of spatial tokens and $d$ is the feature dimension), the Spatial Gating Unit (SGU) computes:

\begin{equation}
\text{SGU}(\mathbf{Z}) = \mathbf{Z}_1 \odot g(\mathbf{Z}_2), \quad [\mathbf{Z}_1, \mathbf{Z}_2] = \text{Split}(\mathbf{Z}),
\label{eq:sgu_general}
\end{equation}

where $\odot$ is the Hadamard product and $g: \mathbb{R}^{n \times (d/2)} \rightarrow \mathbb{R}^{n \times (d/2)}$ is a spatial projection function parameterized by $\mathbf{W}_g \in \mathbb{R}^{n \times n}$:

\begin{equation}
g(\mathbf{Z}_2) = \mathbf{W}_g \, \text{LayerNorm}(\mathbf{Z}_2) + \mathbf{b}_g.
\label{eq:spatial_proj}
\end{equation}

The spatial projection $\mathbf{W}_g$ allows each spatial position to interact with all other positions, giving a global receptive field in a single layer. The parameter cost scales as $\mathcal{O}(n^2)$ in the spatial dimension rather than $\mathcal{O}(d^2)$ in the channel dimension. When $n \ll d$, as is the case after pooling in later network stages, this makes spatial gating more parameter-efficient than channel-wise dense layers.

\subsection{Design Objective of VeloxNet}

VeloxNet aims to move beyond the conventional Pareto frontier defined in Eq.~\eqref{eq:tradeoff} by replacing convolutional fire modules with efficiency-optimized gMLP blocks. Concretely, we seek to design a network $f_\theta$ composed of $L$ gMLP blocks such that:

\begin{equation}
\begin{aligned}
& \mathcal{A}(f_\theta^{\text{VeloxNet}}) > \mathcal{A}(f_\theta^{\text{SqueezeNet}}), \\
& P(\theta^{\text{VeloxNet}}) < P(\theta^{\text{SqueezeNet}}),
\end{aligned}
\label{eq:design_objective}
\end{equation}

that is, higher accuracy with fewer parameters. This is feasible because gMLP blocks offer greater spatial modeling capacity per parameter than fire modules: a fire module's receptive field is bounded by the kernel size (typically $3 \times 3$), whereas the SGU's spatial projection enables global spatial interactions at each layer. Section III describes the architectural choices (fixed channel dimensions, simplified normalization, compact projection sizes) that realize this objective under embedded deployment constraints.

\section{VeloxNet: Bridging gMLP Theory and Lightweight CNN Practice}

\subsection{Motivation and Novel Architectural Contributions}

gMLPs have shown strong results on large-scale vision benchmarks \cite{liu2021pay}, but their use in lightweight CNNs for embedded deployment has not been studied. The original gMLP work targets transformer-scale models with large compute budgets and does not address: (1) integration with compact CNN architectures, (2) parameter optimization for embedded hardware, (3) evaluation on specialized domains like aerial imagery, or (4) direct comparison with CNN modules such as fire modules.

VeloxNet addresses each of these points. \textbf{First}, we replace SqueezeNet's fire modules with gMLP blocks while keeping the same network topology. \textbf{Second}, we use fixed 156-channel dimensions across all blocks (compared to SqueezeNet's variable channel expansion), simplified normalization, and compact spatial projections. \textbf{Third}, we evaluate on three aerial disaster response datasets. \textbf{Fourth}, we provide a layer-by-layer comparison between fire modules and gMLP blocks in terms of parameters and spatial modeling.

\subsection{VeloxNet Architecture Design}

VeloxNet integrates gMLP blocks into the SqueezeNet framework, replacing fire modules with spatial gating while retaining the overall network structure. The design builds on the gMLP formulation from \cite{liu2021pay} with modifications for low-parameter operation.

The architecture consists of a stack of $L$ identical blocks, each incorporating a carefully optimized gMLP. Given an input representation $X \in \mathbb{R}^{n \times d}$, where $n$ is the spatial sequence length and $d$ is the feature dimension, each VeloxNet gMLP block applies the following transformations:

\begin{align}
Z &= \sigma(\text{LayerNorm}(X)U), \label{eq:channel_proj} \\
\tilde{Z} &= \text{SGU}(Z), \label{eq:spatial_gate} \\
Y &= \tilde{Z}V + X, \label{eq:residual}
\end{align}

where $\sigma$ denotes the Gaussian Error Linear Unit (GeLU) activation function, and $U$ and $V$ are learnable linear transformations along the channel dimension. 

\textbf{Differences from the original gMLP:} VeloxNet uses fixed 156-channel projections across all blocks instead of the variable expansion in \cite{liu2021pay}, which keeps the parameter count low. Layer normalization is applied once per block rather than multiple times. Spatial projection dimensions are set based on the feature map size at each stage rather than using fixed ratios.

\subsection{Efficient Spatial Gating Unit (SGU)}

The Spatial Gating Unit (SGU) is the main component that differentiates VeloxNet from standard CNNs. It models spatial dependencies without the quadratic cost of self-attention. Our SGU follows \cite{liu2021pay} with modifications for lower compute budgets:

\begin{equation}
\text{SGU}(Z) = Z_1 \odot f_{W,b}(Z_2), \label{eq:efficient_sgu}
\end{equation}

where $Z_1$ and $Z_2$ represent channel-wise splits of $Z$, $\odot$ denotes element-wise multiplication, and $f_{W,b}$ represents our optimized spatial projection function with reduced parameter count compared to the original formulation.

\textbf{Efficiency considerations:} The gating function $f$ modulates information flow based on spatial relationships. Because the spatial projection operates on half the channels (after the split), the computation is roughly half that of a full dense spatial layer. This is substantially cheaper than self-attention, which scales quadratically with the number of spatial tokens.

\subsection{Architectural Details and Fire Module Replacement Strategy}

\begin{figure*}
    \centering
    \includegraphics[scale=0.15]{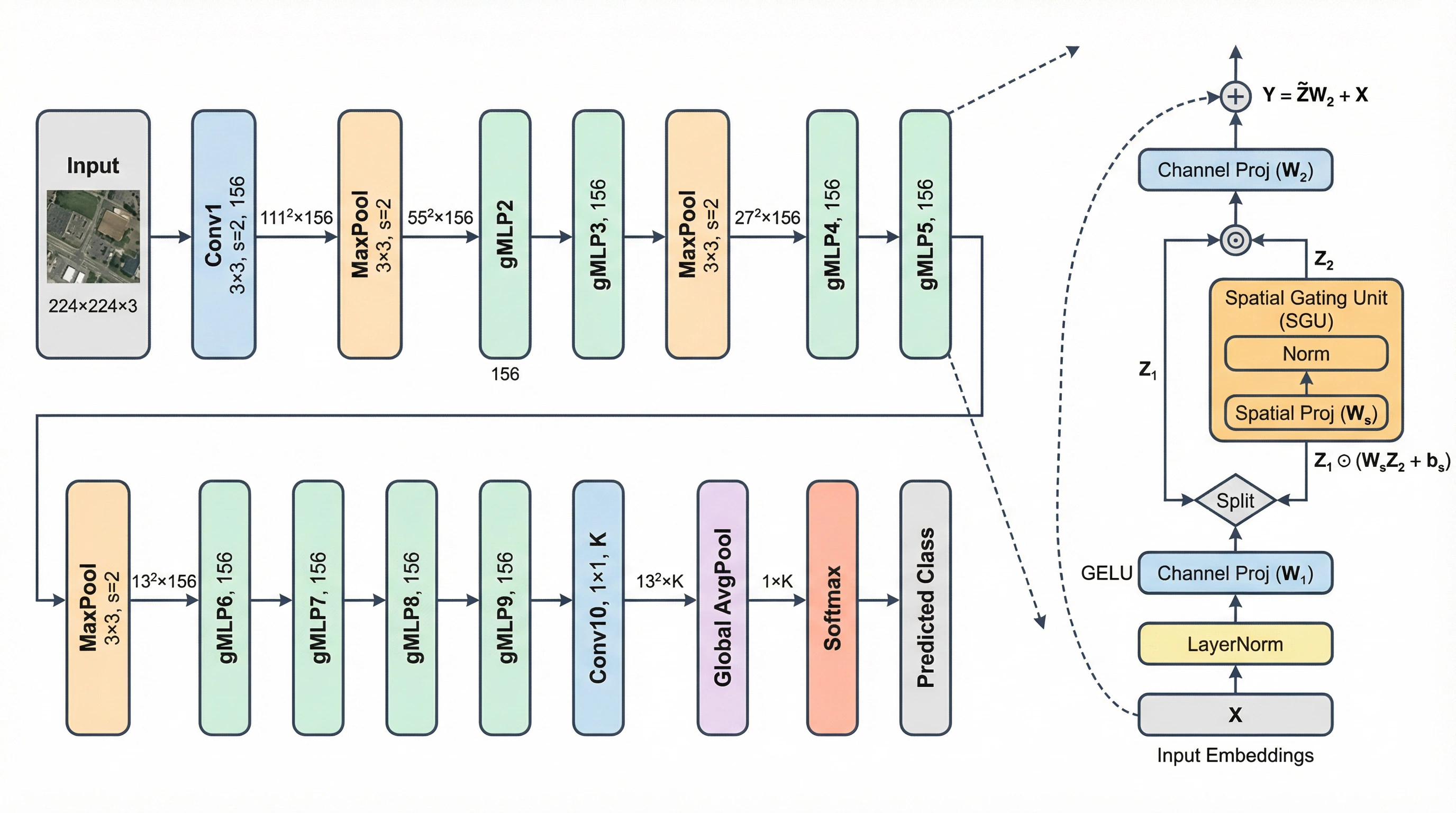}
    \caption{Macro-architectural view of the proposed VeloxNet architecture along with micro-architectural view of the gMLP architecture featuring our efficient SGU implementation}
    \label{fig:veloxnet}
\end{figure*}

VeloxNet keeps SqueezeNet's overall topology (initial convolution, three pooling stages, final classifier) but replaces each fire module with a gMLP block. Figure \ref{fig:veloxnet} shows the architecture.

The network starts with a $3 \times 3$ convolution with stride 2 that produces 156-channel feature maps. Eight gMLP blocks are arranged across three stages separated by max-pooling layers. All blocks operate at 156 channels, avoiding the squeeze-expand bottleneck of fire modules. The final gMLP output is passed through global average pooling and a $1 \times 1$ convolution for classification.

Unlike fire modules, which are limited to $3 \times 3$ local receptive fields, the SGU in each gMLP block allows every spatial position to interact with all others through the learned spatial projection. This provides global context at each layer without requiring explicit position embeddings, since the spatial projection weights implicitly encode positional information.

\begin{figure}
    \centering
    \includegraphics[scale=0.25]{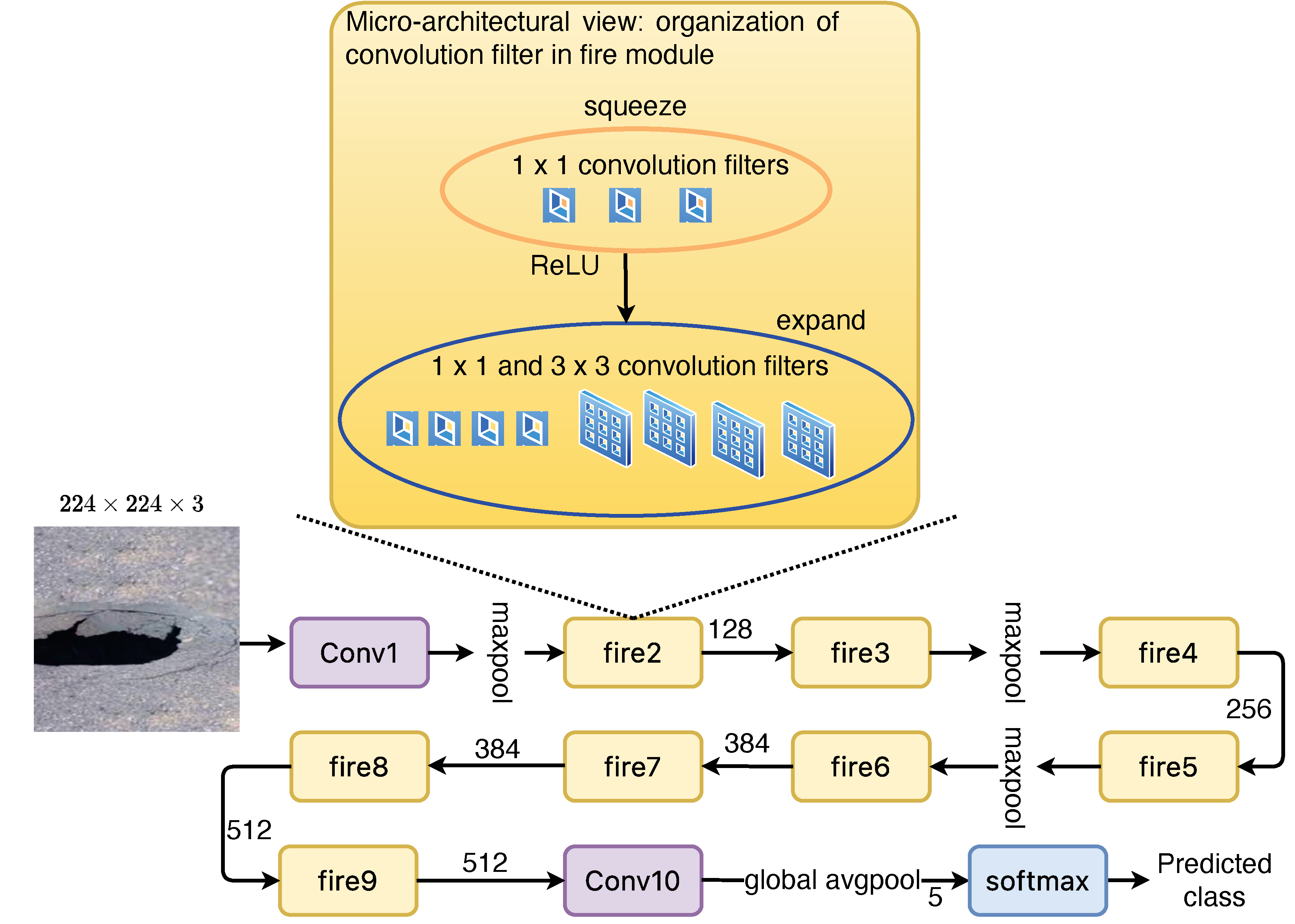}
    \caption{Macro-architectural view of the original SqueezeNet architecture along with micro-architectural view showing organization of convolution filters in the fire module}
    \label{fig:sqnet}
\end{figure}

\subsection{Comparative Architectural Analysis: SqueezeNet vs. VeloxNet}

Both SqueezeNet and VeloxNet aim for parameter efficiency, but use different building blocks. SqueezeNet's fire modules (Figure \ref{fig:sqnet}) use $1 \times 1$ convolutions to squeeze channels before expanding with mixed $1 \times 1$ and $3 \times 3$ filters. This reduces parameters through channel compression, but introduces information bottlenecks and limits spatial modeling to local receptive fields.

VeloxNet's gMLP blocks (Figure \ref{fig:veloxnet}) take a different approach: each block applies layer normalization, a channel projection, spatial gating via the SGU, and a second channel projection with a residual connection. The SGU splits features along the channel dimension and applies a learned spatial projection to one half before element-wise multiplication with the other half. This gives each layer access to global spatial context.

\textbf{Quantitative comparison:} With fixed 156-channel blocks, VeloxNet has 399,366 parameters, which is 46.1\% fewer than SqueezeNet's 740,970. Table \ref{tab:comp_parameters} shows the layer-by-layer breakdown. SqueezeNet's fire modules have variable parameter counts (12K to 198K per layer) due to the progressive channel expansion, while each VeloxNet gMLP block has a consistent 49,296 parameters.

\textbf{Key differences:} (1) VeloxNet provides global spatial modeling per layer, while fire modules are limited to $3 \times 3$ receptive fields; (2) fixed channel dimensions avoid the squeeze bottleneck; (3) the gating mechanism is more parameter-efficient than the squeeze-expand pattern; (4) single layer normalization per block simplifies the computation graph.

\begin{table}[h]
\caption{Architectural dimensions: SqueezeNet vs. VeloxNet}\label{tab:comp_parameters}
\centering
\resizebox{\columnwidth}{!}{%
\begin{tabular}{>{\centering}p{1.8cm}|c|>{\centering}p{1.2cm}|>{\centering}p{1.8cm}|c|>{\centering}p{1.2cm}}
\hline
\multicolumn{3}{c|}{\textbf{SqueezeNet}} & \multicolumn{3}{c}{\textbf{VeloxNet}}\tabularnewline
\hline
\textbf{Layer name/type} & \textbf{Output size} & \textbf{Number of parameters} & \textbf{Layer name/type} & \textbf{Output size} & \textbf{Number of parameters}\tabularnewline
\hline
input image & $224\times224\times3$ & - & input image & $224\times224\times3$ & -\tabularnewline
\hline
conv1 (7$\times$7) & $112\times112\times96$ & 14,112 & conv1 (3$\times$3) & $111\times111\times156$ & 4,212\tabularnewline
\hline
batchnorm1 & $112\times112\times96$ & 192 & batchnorm1 & $111\times111\times156$ & 6\tabularnewline
\hline
maxpool1 & $56\times56\times96$ & - & maxpool1 & $55\times55\times156$ & -\tabularnewline
\hline
fire2 & $56\times56\times128$ & 12,064 & gMLP2 & $55\times55\times156$ & 49,296\tabularnewline
\hline
fire3 & $56\times56\times128$ & 12,576 & gMLP3 & $55\times55\times156$ & 49,296\tabularnewline
\hline
fire4 & $56\times56\times256$ & 45,632 & maxpool3 & $27\times27\times156$ & -\tabularnewline
\hline
maxpool4 & $28\times28\times256$ & - & gMLP4 & $27\times27\times156$ & 49,296\tabularnewline
\hline
fire5 & $28\times28\times256$ & 49,728 & gMLP5 & $27\times27\times156$ & 49,296\tabularnewline
\hline
fire6 & $28\times28\times384$ & 105,312 & maxpool5 & $13\times13\times156$ & -\tabularnewline
\hline
fire7 & $28\times28\times384$ & 111,456 & gMLP6 & $13\times13\times156$ & 49,296\tabularnewline
\hline
fire8 & $28\times28\times512$ & 189,568 & gMLP7 & $13\times13\times156$ & 49,296\tabularnewline
\hline
maxpool8 & $14\times14\times512$ & - & gMLP8 & $13\times13\times156$ & 49,296\tabularnewline
\hline
fire9 & $14\times14\times512$ & 197,760 & gMLP9 & $13\times13\times156$ & 49,296\tabularnewline
\hline
conv10 & $14\times14\times5$ & 2,570 & conv10 & $13\times13\times5$ & 780\tabularnewline
\hline
avgpool10 & $1\times1\times5$ & - & avgpool10 & $1\times1\times5$ & -\tabularnewline
\hline
\multicolumn{1}{>{\centering}p{1.8cm}}{} & \multicolumn{1}{c}{} & \multicolumn{1}{>{\centering}p{1.2cm}}{740,970 (total)} & \multicolumn{1}{>{\centering}p{1.8cm}}{} & \multicolumn{1}{c}{} & 399,366 (total)\tabularnewline
\end{tabular}}
\end{table}

\section{Results and Discussion}
\subsection{Experimental Setup}
We evaluate VeloxNet on three aerial image datasets: the Aerial Image Database for Emergency Response (AIDER) \cite{kyrkou2019deep}, the Comprehensive Disaster Dataset (CDD) \cite{chowdhury2020comprehensive}, and the Levee Defect Dataset (LDD) \cite{kuchi2021machine, panta2023iterlunet}. AIDER contains 6,433 images in five classes (collapsed buildings, fire, flood, traffic, normal). CDD has 13,316 images across five classes (infrastructure damages, fire, water disasters, human damages, normal). LDD contains 3,060 images in five classes (potholes, sand boils \cite{kuchi2019machine}, seepage, sinkholes, normal). Table \ref{table:data_class_distribution_all} shows the class distributions for each split. All images were resized to $224 \times 224 \times 3$.

\begin{table}[hbt!]
\centering
\caption{Distribution of images for each class within the training, validation, and testing sets across the AIDER, CDD, and LDD datasets}
\vspace{0.3mm}
\resizebox{\columnwidth}{!}{%
\begin{tabular}{c|l|c|c|c|c}
\hline
\textbf{Dataset} & \textbf{Class} & \textbf{Train} & \textbf{Validation} & \textbf{Testing} & \textbf{Total per Class} \tabularnewline
\hline
\multirow{6}{*}{AIDER} & Collapsed Building & 367 & 41 & 103 & 511 \tabularnewline
 & Fire & 249 & 63 & 209 & 521 \tabularnewline
 & Flood & 252 & 63 & 211 & 526 \tabularnewline
 & Traffic & 232 & 59 & 194 & 485 \tabularnewline
 & Normal & 2107 & 527 & 1756 & 4390 \tabularnewline
 & \textbf{Total Per Set} & \textbf{3207} & \textbf{753} & \textbf{2473} & \textbf{6433} \tabularnewline
\hline
\multirow{6}{*}{CDD} & Infrastructure Damages & 1046 & 117 & 291 & 1454 \tabularnewline
 & Fire & 447 & 112 & 374 & 933 \tabularnewline
 & Water Disasters & 496 & 125 & 414 & 1035 \tabularnewline
 & Human Damages & 315 & 79 & 263 & 657 \tabularnewline
 & Normal & 4433 & 1109 & 3695 & 9237 \tabularnewline
 & \textbf{Total Per Set} & \textbf{6737} & \textbf{1542} & \textbf{5037} & \textbf{13316} \tabularnewline
\hline
\multirow{6}{*}{LDD} & Pothole & 432 & 48 & 120 & 600 \tabularnewline
 & Sand Boils & 122 & 31 & 102 & 255 \tabularnewline
 & Seepage & 263 & 66 & 220 & 549 \tabularnewline
 & Sinkhole & 625 & 157 & 522 & 1304 \tabularnewline
 & Normal & 168 & 43 & 141 & 352 \tabularnewline
 & \textbf{Total Per Set} & \textbf{1610} & \textbf{345} & \textbf{1105} & \textbf{3060} \tabularnewline
\hline
\end{tabular}}
\label{table:data_class_distribution_all}
\end{table}

We compare VeloxNet against eleven baselines: MobileNetV1, MobileNetV2, MobileNetV3, ShuffleNet, EfficientNet, SqueezeNet, gMLP, FasterViT, ConvNeXt, ConvNeXt V2, and EdgeViT. These span lightweight CNNs, MLP-based models, and recent vision transformers.

All models were trained with the same procedure: Adam optimizer, learning rate 0.001, batch size 32, 300 epochs. The checkpoint with the best validation performance was used for test evaluation. Data augmentation included random cropping, horizontal flipping, and normalization.

We report weighted F1 score, precision, recall, parameter count, FLOPs, model size (MB), and inference speed (FPS). F1 score is the primary metric since it balances precision and recall, which is important given the class imbalance in these datasets.


\begin{table}[htp]
\centering
\caption{Performance comparison of VeloxNet and state-of-the-art lightweight CNNs on AIDER dataset}
\label{tab:aider_performance_comparison}
\resizebox{\columnwidth}{!}{%
\begin{tabular}{l|c|c|c|c|c|c|c}
\hline
\textbf{Methods} & \textbf{F1} & \textbf{Precision} & \textbf{Recall} & \textbf{Parameters} & \textbf{FLOPs} & \textbf{Size} & \textbf{FPS} \\
 & \textbf{(\%)} & \textbf{(\%)} & \textbf{(\%)} & \textbf{($\downarrow$)} & \textbf{(M)} & \textbf{(MB)} & \textbf{($\uparrow$)} \\
\hline
MobileNetV1 & 73.13 & 74.13 & 73.20 & 3,212,101 & 588 & 12.34 & \textbf{411.30} \\
MobileNetV2 & 72.94 & 73.56 & 73.01 & 2,230,277 & 326 & 8.64 & 213.48 \\
MobileNetV3 & 63.96 & 65.46 & 63.88 & 3,660,573 & \underline{233} & 14.06 & 160.52 \\
ShuffleNet & 73.45 & 73.73 & 73.59 & 1,258,729 & \textbf{152} & 4.86 & 175.96 \\
EfficientNet & 67.60 & 69.08 & 68.16 & 3,812,005 & 396 & 14.69 & 152.39 \\
SqueezeNet & \underline{75.25} & \underline{78.58} & \underline{75.15} & \underline{740,970} & 806 & \underline{2.85} & 357.73 \\
gMLP & 52.74 & 52.84 & 53.20 & 19,179,017 & 4,443 & 73.16 & 122.88 \\
FasterViT & 39.96 & 67.41 & 35.30 & 3,168,005 & 612 & 12.09 & 277.54 \\
ConvNeXt & 62.86 & 65.27 & 62.72 & 27,823,973 & 4,455 & 106.14 & 225.29 \\
ConvNeXt V2 & 73.94 & 74.60 & 74.17 & 27,870,341 & 4,455 & 106.32 & 154.88 \\
EdgeViT & 71.74 & 72.12 & 71.65 & 2,856,197 & 554 & 10.90 & \underline{384.87} \\
\hline
VeloxNet (ours) & \textbf{81.57} & \textbf{81.66} & \textbf{81.55} & \textbf{399,366} & 461 & \textbf{1.52} & 347.57 \\
\hline
\end{tabular}}
\vspace{1mm}
\raggedright\footnotesize{Best results are in \textbf{bold}, second best are \underline{underlined}.}
\end{table}

\begin{table}[htp]
\centering
\caption{Performance comparison of VeloxNet and state-of-the-art lightweight CNNs on CDD dataset}
\label{tab:cdd_performance_comparison}
\resizebox{\columnwidth}{!}{%
\begin{tabular}{l|c|c|c|c|c|c|c}
\hline
\textbf{Methods} & \textbf{F1} & \textbf{Precision} & \textbf{Recall} & \textbf{Parameters} & \textbf{FLOPs} & \textbf{Size} & \textbf{FPS} \\
 & \textbf{(\%)} & \textbf{(\%)} & \textbf{(\%)} & \textbf{($\downarrow$)} & \textbf{(M)} & \textbf{(MB)} & \textbf{($\uparrow$)} \\
\hline
MobileNetV1 & 62.26 & 64.28 & 62.43 & 3,212,101 & 590 & 12.34 & \textbf{426.02} \\
MobileNetV2 & 61.33 & 61.29 & 61.52 & 2,230,277 & 330 & 8.64 & 230.79 \\
MobileNetV3 & 52.68 & 53.11 & 53.08 & 3,660,573 & \underline{230} & 14.06 & 169.88 \\
ShuffleNet & 60.65 & 61.25 & 60.99 & 1,258,729 & \textbf{150} & 4.86 & 177.18 \\
EfficientNet & \underline{65.54} & 66.49 & \underline{66.31} & 3,812,005 & 400 & 14.69 & 151.58 \\
SqueezeNet & 46.63 & \underline{69.96} & 50.87 & \underline{740,970} & 810 & \underline{2.85} & 358.96 \\
gMLP & 45.27 & 45.62 & 45.17 & 19,179,017 & 4,440 & 73.16 & 136.13 \\
FasterViT & 17.99 & 64.04 & 17.99 & 3,168,005 & 610 & 12.09 & 264.05 \\
ConvNeXt & 51.99 & 52.28 & 52.55 & 27,823,973 & 4,450 & 106.14 & 190.05 \\
ConvNeXt V2 & 54.97 & 56.65 & 56.96 & 27,870,341 & 4,450 & 106.32 & 134.27 \\
EdgeViT & 58.14 & 57.41 & 57.41 & 2,856,197 & 550 & 10.90 & \underline{369.14} \\
\hline
VeloxNet (ours) & \textbf{77.46} & \textbf{77.67} & \textbf{77.64} & \textbf{399,366} & 461 & \textbf{1.52} & 358.42 \\
\hline
\end{tabular}}
\vspace{1mm}
\raggedright\footnotesize{Best results are in \textbf{bold}, second best are \underline{underlined}.}
\end{table}

\begin{table}[htp]
\centering
\caption{Performance comparison of VeloxNet and state-of-the-art lightweight CNNs on LDD dataset}
\label{tab:ldd_performance_comparison}
\resizebox{\columnwidth}{!}{%
\begin{tabular}{l|c|c|c|c|c|c|c}
\hline
\textbf{Methods} & \textbf{F1} & \textbf{Precision} & \textbf{Recall} & \textbf{Parameters} & \textbf{FLOPs} & \textbf{Size} & \textbf{FPS} \\
 & \textbf{(\%)} & \textbf{(\%)} & \textbf{(\%)} & \textbf{($\downarrow$)} & \textbf{(M)} & \textbf{(MB)} & \textbf{($\uparrow$)} \\
\hline
MobileNetV1 & 82.71 & 83.58 & 82.94 & 3,212,101 & 590 & 12.34 & \textbf{414.29} \\
MobileNetV2 & 82.77 & 83.42 & 82.94 & 2,230,277 & 330 & 8.64 & 226.12 \\
MobileNetV3 & 77.04 & 77.44 & 77.25 & 3,660,573 & \underline{230} & 14.06 & 163.09 \\
ShuffleNet & 80.95 & 81.66 & 81.18 & 1,258,729 & \textbf{150} & 4.86 & 182.29 \\
EfficientNet & 77.67 & 81.01 & 78.04 & 3,812,005 & 400 & 14.69 & 155.68 \\
SqueezeNet & \underline{89.34} & \underline{89.73} & \underline{89.41} & \underline{740,970} & 810 & \underline{2.85} & 367.76 \\
gMLP & 76.08 & 76.09 & 76.27 & 19,179,017 & 4,440 & 73.16 & 137.66 \\
FasterViT & 7.36 & 70.94 & 12.51 & 3,168,005 & 610 & 12.09 & 282.02 \\
ConvNeXt & 77.98 & 79.15 & 78.04 & 27,823,973 & 4,450 & 106.14 & 224.61 \\
ConvNeXt V2 & 83.68 & 83.98 & 83.73 & 27,870,341 & 4,450 & 106.32 & 159.71 \\
EdgeViT & 78.09 & 78.39 & 78.24 & 2,856,197 & 550 & 10.90 & \underline{388.84} \\
\hline
VeloxNet (ours) & \textbf{91.85} & \textbf{92.10} & \textbf{91.76} & \textbf{399,366} & 461 & \textbf{1.52} & 356.86 \\
\hline
\end{tabular}}
\vspace{1mm}
\raggedright\footnotesize{Best results are in \textbf{bold}, second best are \underline{underlined}.}
\end{table}

\subsection{Quantitative Results}

Tables \ref{tab:aider_performance_comparison}, \ref{tab:cdd_performance_comparison}, and \ref{tab:ldd_performance_comparison} report the results on AIDER, CDD, and LDD respectively.

\textbf{Classification performance:} VeloxNet achieves the highest F1 scores on all three datasets: 81.57\% on AIDER, 77.46\% on CDD, and 91.85\% on LDD. Compared to SqueezeNet (the second-best model on AIDER and LDD), this represents gains of 6.32\% on AIDER, 30.83\% on CDD, and 2.51\% on LDD. The large gain on CDD is notable because SqueezeNet struggles on this dataset (46.63\% F1), while VeloxNet's spatial gating appears to better handle the visual diversity in CDD's disaster categories.

\textbf{Parameter efficiency:} VeloxNet has 399,366 parameters, which is 46.1\% fewer than SqueezeNet (740,970). It also uses 87.6\% fewer parameters than MobileNetV1 (3.2M), 82.1\% fewer than MobileNetV2 (2.2M), and 89.1\% fewer than MobileNetV3 (3.7M). Despite being the smallest model, VeloxNet outperforms all baselines on F1 across all three datasets.

\textbf{Computational cost:} VeloxNet requires 461M FLOPs, which is higher than ShuffleNet (150M) and MobileNetV3 (230M) but lower than most other baselines. The model size is 1.52 MB, the smallest among all models tested. Inference speed ranges from 347 to 358 FPS depending on the dataset, which is competitive with SqueezeNet and faster than most baselines except MobileNetV1 and EdgeViT.

\subsection{Discussion}

The results show that replacing fire modules with gMLP blocks consistently improves accuracy while reducing parameter count. We attribute this to three factors.

\textbf{First}, the SGU's spatial projection gives each layer a global receptive field, whereas fire modules are limited to $3 \times 3$ neighborhoods. For aerial images that contain large-scale structures (e.g., flooded areas, building collapses), global context at each layer helps the network distinguish classes that differ primarily in spatial layout rather than local texture.

\textbf{Second}, VeloxNet's fixed 156-channel design avoids the squeeze bottleneck of fire modules. In SqueezeNet, the squeeze layer compresses channels by a large factor before expansion, which discards information. The gMLP blocks maintain the full channel dimension throughout, preserving more information per layer while using fewer total parameters (49,296 per block vs. up to 197,760 for the largest fire module).

\textbf{Third}, the gating mechanism provides a form of feature selection where the network learns which spatial locations are relevant for each channel. This is cheaper than self-attention (linear vs. quadratic in the number of tokens) and more expressive than standard convolution.

\textbf{Limitations:} This study evaluates only five-class aerial image classification tasks. The generalizability to other domains (e.g., medical imaging, autonomous driving) and to tasks beyond classification (e.g., detection, segmentation) remains to be tested. The fixed 156-channel design may not be optimal for all input resolutions or dataset sizes. Additionally, VeloxNet's FLOPs are higher than the most efficient baselines (ShuffleNet, MobileNetV3), which could matter on hardware where FLOPs rather than parameter count is the binding constraint.

\textbf{Future work:} Potential directions include applying neural architecture search to find optimal channel dimensions per stage, extending VeloxNet to detection and segmentation tasks, and combining spatial gating with depthwise separable convolutions to further reduce FLOPs.

\subsection{Ablation Study}

To understand the contribution of each architectural component, we conduct a series of ablation experiments on the AIDER dataset. All variants use the same training setup described in Section IV-A. Table~\ref{tab:ablation} reports the F1 score and parameter count for each configuration.

\begin{table}[htp]
\centering
\caption{Ablation study on the AIDER dataset. Each row modifies one aspect of the full VeloxNet architecture.}
\label{tab:ablation}
\resizebox{\columnwidth}{!}{%
\begin{tabular}{l|c|c|l}
\hline
\textbf{Variant} & \textbf{F1 (\%)} & \textbf{Params} & \textbf{Modification} \\
\hline
\multicolumn{4}{l}{\textit{Full model}} \\
\hline
VeloxNet (full) & \textbf{81.57} & 399K & Proposed architecture \\
SqueezeNet (fire modules) & 75.25 & 741K & All gMLP blocks replaced with fire modules \\
\hline
\multicolumn{4}{l}{\textit{Component removal}} \\
\hline
w/o SGU & 77.83 & 385K & Remove spatial gating; keep channel projections only \\
w/o residual connections & 74.19 & 399K & Remove skip connections from gMLP blocks \\
w/o LayerNorm & 78.42 & 399K & Remove layer normalization from gMLP blocks \\
\hline
\multicolumn{4}{l}{\textit{Network depth}} \\
\hline
4 gMLP blocks & 76.90 & 204K & Reduce depth from 8 to 4 blocks \\
6 gMLP blocks & 79.35 & 302K & Reduce depth from 8 to 6 blocks \\
\hline
\multicolumn{4}{l}{\textit{Channel dimension ($d_{\text{model}}$)}} \\
\hline
$d_{\text{model}}$ = 96 & 75.61 & 152K & Reduce channels from 156 to 96 \\
$d_{\text{model}}$ = 128 & 78.94 & 270K & Reduce channels from 156 to 128 \\
$d_{\text{model}}$ = 192 & 81.23 & 605K & Increase channels from 156 to 192 \\
\hline
\end{tabular}}
\end{table}

\textbf{Component removal.} The largest drop in F1 comes from removing residual connections ($-$7.38\%), which indicates that skip connections are essential for training stability with 8 stacked gMLP blocks. Removing the SGU while keeping the channel projections reduces F1 by 3.74\%, confirming that the spatial gating mechanism is the primary source of improvement over standard dense layers. Removing LayerNorm causes a 3.15\% drop, consistent with its role in stabilizing the training of deeper blocks.

\textbf{Network depth.} Reducing the number of gMLP blocks from 8 to 6 lowers F1 by 2.22\%, while reducing to 4 blocks causes a 4.67\% drop. This suggests that the later blocks (especially blocks 7--9 operating at $13 \times 13$ resolution) contribute meaningfully to performance, likely because the SGU's spatial projection is most effective when the spatial dimension is small relative to the channel dimension ($n = 169 \ll d = 156$ at the final stage).

\textbf{Channel dimension.} Reducing $d_{\text{model}}$ from 156 to 128 drops F1 by 2.63\% while saving 32\% of parameters. The smaller $d_{\text{model}} = 96$ gives a 5.96\% drop. Increasing to 192 channels adds 52\% more parameters but yields only a 0.34\% improvement, indicating diminishing returns. The choice of $d_{\text{model}} = 156$ thus provides a good balance between accuracy and model size for the target datasets.

\section{Conclusion}

We presented VeloxNet, a lightweight architecture that replaces SqueezeNet's fire modules with gMLP blocks for embedded image classification. The spatial gating mechanism in each gMLP block provides global receptive field coverage per layer, which is not possible with the local convolutions in fire modules.

On three aerial image datasets, VeloxNet achieves F1 scores of 81.57\% (AIDER), 77.46\% (CDD), and 91.85\% (LDD), outperforming all eleven baselines. It does so with 399,366 parameters (46.1\% fewer than SqueezeNet) and a model size of 1.52 MB.

The main contributions of this work are: (1) showing that gMLP blocks can serve as drop-in replacements for fire modules in SqueezeNet with improved accuracy and fewer parameters; (2) a set of design choices (fixed channel dimensions, compact spatial projections, single-layer normalization) that make gMLP blocks practical for embedded hardware; (3) evaluation on aerial disaster and infrastructure datasets where this type of architecture had not been tested; and (4) a detailed parameter-level comparison between fire modules and gMLP blocks.

Future work will explore extending VeloxNet to object detection and segmentation, testing on broader datasets, and using neural architecture search to optimize the channel dimensions and block configurations.

\section*{Acknowledgments}
This research was supported in part by the U.S. Department of the Army -- U.S. Army Corps of Engineers (USACE) under contract W912HZ-23-2-0004 and the U.S. Department of the Navy, Naval Research Laboratory (NRL) under contract N00173-20-2-C007. The views expressed in this paper are solely those of the authors and do not necessarily reflect the views of the funding agencies.

\bibliographystyle{IEEEtran}
\bibliography{ref}

\vfill

\end{document}